\documentclass[letterpaper]{article} 
\usepackage[]{aaai25}  
\usepackage{times}  
\usepackage{helvet}  
\usepackage{courier}  
\usepackage[hyphens]{url}  
\usepackage{graphicx} 
\usepackage{natbib}  
\usepackage{caption} 
\usepackage{algorithm}
\usepackage{algorithmic}
\usepackage[dvipsnames]{xcolor}

\usepackage{newfloat}
\usepackage{listings}
\DeclareCaptionStyle{ruled}{labelfont=normalfont,labelsep=colon,strut=off} 
\floatstyle{ruled}
\newfloat{listing}{tb}{lst}{}
\floatname{listing}{Listing}
\title{\textsc{SummPilot}: Bridging Efficiency and Customization \\ for Interactive Summarization System}
\author{
    JungMin Yun\textsuperscript{\rm 1},
    Juhwan Choi\textsuperscript{\rm 1},
    Kyohoon Jin\textsuperscript{\rm 2},
    Soojin Jang\textsuperscript{\rm 2},
    Jinhee Jang\textsuperscript{\rm 1},
    YoungBin Kim\textsuperscript{\rm 1,2}
}
\affiliations{
    \textsuperscript{\rm 1}Department of Artificial Intelligence, Chung-Ang University\\
    \textsuperscript{\rm 2}Graduate School of Advanced Imaging Sciences, Multimedia and Film, Chung-Ang University\\
    
    \{cocoro357, gold5230, fhzh123, sujin0110, jinheejang, ybkim85\}@cau.ac.kr
}

\usepackage{bibentry}

\begin{document}

\maketitle

\begin{abstract}
This paper incorporates the efficiency of automatic summarization and addresses the challenge of generating personalized summaries tailored to individual users' interests and requirements. To tackle this challenge, we introduce \textsc{SummPilot}, an interaction-based customizable summarization system. \textsc{SummPilot} leverages a large language model to facilitate both automatic and interactive summarization. Users can engage with the system to understand document content and personalize summaries through interactive components such as semantic graphs, entity clustering, and explainable evaluation. Our demo and user studies demonstrate \textsc{SummPilot}'s adaptability and usefulness for customizable summarization.
\end{abstract}

\section{Introduction \& Related Work}

Advancements in deep learning have greatly enhanced summarization systems, but automated systems often fall short of addressing individual requirements or interests. Interactive systems address this by enabling personalized summaries through user engagement~\cite{hirsch-etal-2021-ifacetsum, adenuga2022living, slobodkin-etal-2023-summhelper, zhang2023concepteva}. 

However, existing interactive summarization systems struggle to represent content relationships, which are essential for understanding connections within or across documents. This hampers multi-document summarization, where users need to integrate information from various sources. Furthermore, these systems lack support for decision-making by failing to indicate how user inputs affect the final summary, making it difficult for users to assess the impact of their choices and reducing the effectiveness of interactions.

To address this limitation, we adopt an interactive, human-in-the-loop approach that incorporates user-specific needs into the summarization process. With this approach, we aim to achieve customizable and controllable summaries that align with diverse and personalized requirements. In this context, we introduce \textsc{SummPilot}, an interactive summarization system that leverages the flexibility and advanced language understanding capabilities of large language models (LLMs). To effectively handle lengthy or multi-topic documents—where customizable summarization becomes crucial—\textsc{SummPilot} also supports multi-document summarization through graph-based connections that capture relationships between key pieces of information.

\begin{figure}[t]
\centering
\includegraphics[width=1.0\columnwidth]{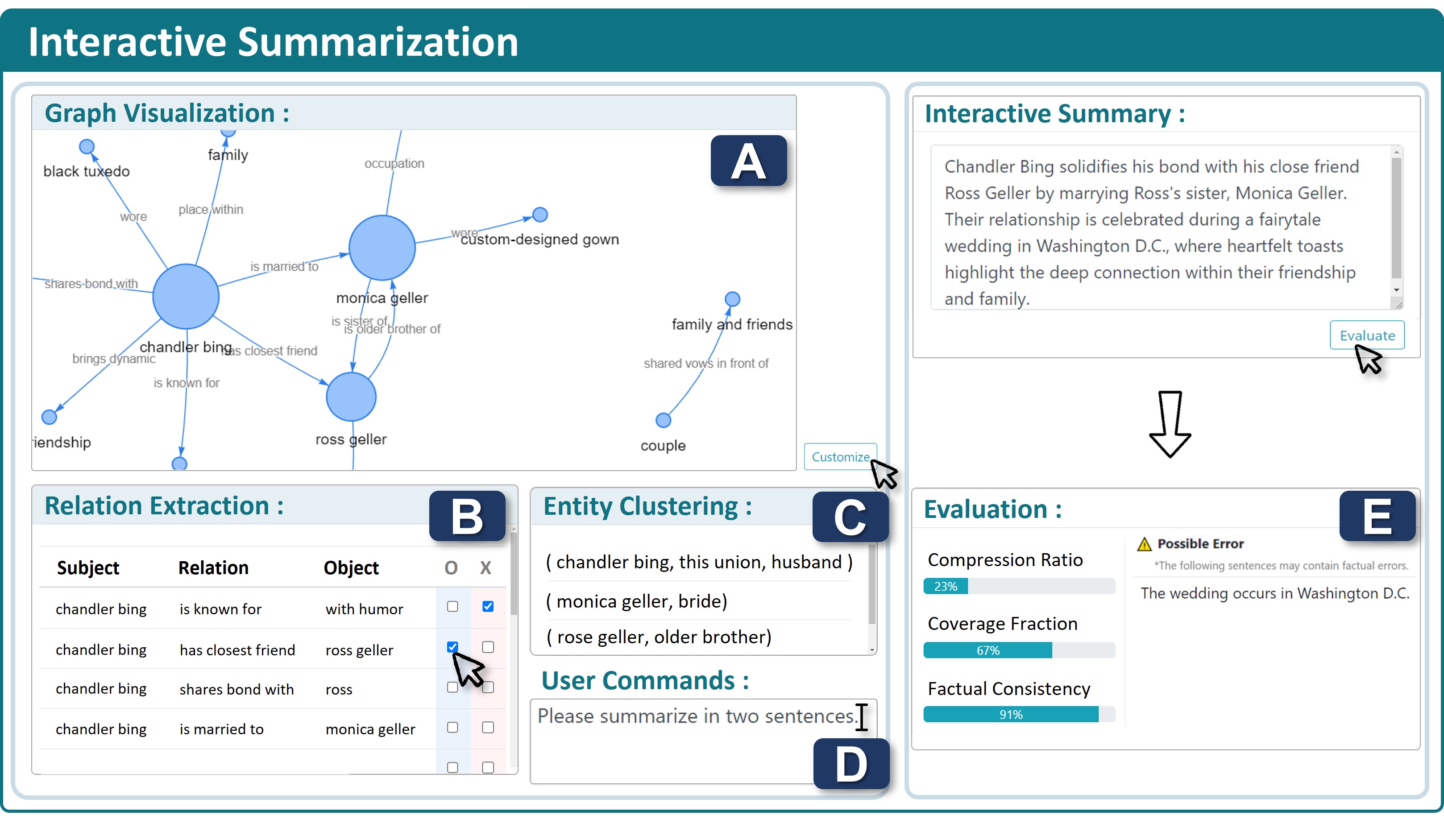}
\caption{Interface design and components for \textsc{SummPilot} (\textsc{Advanced Mode}): Users can explore documents through semantic graphs based on relational triples [\texttt{A}], and manage content via checkboxes for specific triples [\texttt{B}]. Entity clustering groups related expressions [\texttt{C}]. Open-form commands allow custom inputs [\texttt{D}] and the ``Evaluate'' button provides feedback on compression, coverage, and consistency, helping to refine user control and model performance [\texttt{E}].}
\label{fig1}
\end{figure}

\textsc{SummPilot} offers two modes tailored to different summarization needs. \textsc{Basic Mode} delivers automatic summarization by focusing on key information, while \textsc{Advanced Mode} offers user-centric control over the interactive summarization process. This \textsc{Advanced Mode} allows users to specify their information needs more concretely and adjust the content and style of the summary to their preferences, resulting in a more personalized and interactive experience.

\begin{figure*}[t]
\centering
\includegraphics[width=0.9\textwidth]{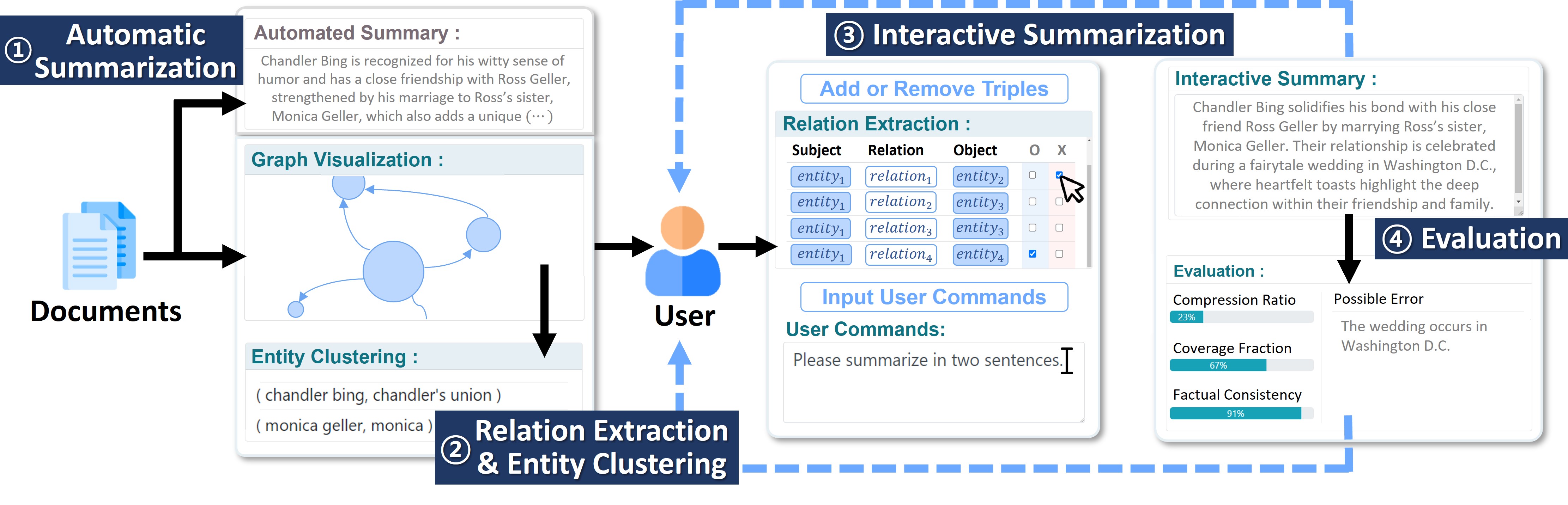}
\caption{Backend pipeline of \textsc{SummPilot}.}
\label{figure_2}
\end{figure*}

Moreover, \textsc{SummPilot} integrates an evaluation mechanism to refine generated summaries by ensuring factual consistency and detecting errors caused by user interventions or model inaccuracies. In \textsc{Advanced Mode}, users receive explainable evaluations that highlight statements with potential factual inaccuracies. This allows users quickly identify possible errors and understand how their input influences the summary, facilitating efficient correction and refinement. By fostering collaborative improvement, this evaluation system establishes a feedback loop between the user and the model, ultimately enhancing the summarization process.

\section{Method: \textsc{SummPilot}}
As shown in Figure~\ref{figure_2}, \textsc{SummPilot} employs multiple internal modules to generate and customize summaries. By leveraging LLMs (specifically GPT-4o \cite{openai2024gpt4o}), our system achieves significant flexibility through well-designed prompts and advanced language comprehension. 

\subsubsection{Automatic Summarization} When multiple documents are provided as input, the automatic summarization module utilizes an LLM to generate a coherent abstractive summary.

\subsubsection{Relation Extraction and Entity Clustering} The relation extraction module identifies relational triples from input documents, forming the basis of a semantic graph that visually represents the information. \textsc{SummPilot}'s graph visualization maps these triples with directed edges to indicate relationships and adjusts node sizes to emphasize key entities. This helps users better understand the content structure and precisely control their information needs. After generating the graph, entity clustering groups mentions of the same entities. For example, the triples $<$Tom-is married to-Jane$>$ and $<$Tom's wife-aged-30$>$ are combined into $<$[Tom's wife, Jane]-aged-30$>$. This clustering clarifies variations in expressions referring to the same concept. A representative entity, determined by frequency, is then assigned to each subject and object for visualization and interaction.

\subsubsection{Interactive Summarization} The interactive summarization module generates customized summaries based on user needs. This module interprets user requests and incorporates them into prompts for the LLM, addressing inclusion/exclusion criteria and user-defined preferences in both structured and flexible contexts. This adaptability supports iterative refinement, enhancing user engagement and satisfaction.

\subsubsection{Summary Evaluation} The evaluation of generated summaries involves three components: compression, coverage, and factual consistency. Compression measures how much the summary condenses the original text, while coverage assess how well the summary retains information from the source documents\cite{grusky-etal-2018-newsroom}. Factual consistency, adapted from FactScore \cite{min-etal-2023-factscore} and customized for our task, evaluates the accuracy by decomposing each summary into atomic facts and verifying them against the original documents. The final factual consistency score is the proportion of verified atomic facts. Unverified facts are flagged as ``possible errors" in the interface.

\section{User Studies \& Evaluations}

\begin{table}[t]
\centering
\resizebox{0.8\columnwidth}{!}{%
\begin{tabular}{lc}
\hline
\textbf{Aspect}                                  & \textbf{Score} \\ \hline\hline
Semantic Graph Effectiveness            & 4.4 (0.8)      \\
Entity Clustering Effectiveness         & 3.8 (0.7)      \\
Intuitiveness Relation Extraction       & 5.0 (0.0)      \\
Ease of Control                         & 4.6 (0.5)      \\
Possible Error                          & 4.6 (0.5)      \\
\hline
Effectiveness of Including Requirements & 4.8 (0.4)      \\
Effectiveness of Excluding Requirements & 4.8 (0.4)      \\
Summary Satisfaction          & 4.8 (0.4)      \\
Summary Trustworthiness              & 4.4 (0.8)      \\ \hline
\end{tabular}
}
\caption{Result of usefulness questionnaire. The scores are reported as the average (standard deviation).}
\label{tab-custom}
\end{table}

The results in Table~\ref{tab-custom} highlight several key advantages of \textsc{SummPilot}. Participants appreciated the semantic graph for enhancing document understanding and praised the intuitive interaction enabled by relation extraction. The system's error detection feature was also recognized as highly beneficial for evaluating generated summaries. These findings underscore \textsc{SummPilot}'s strengths in facilitating comprehensive document analysis, enabling precise user interaction, and supporting effective summarization and evaluation.

\section{Conclusions \& Future Work}

\textsc{SummPilot} is an interaction-driven, customizable summarization system designed for flexibility and effectiveness, as demonstrated by user studies. It establishes a robust foundation for efficiently summarizing complex business documents and news articles, making it particularly beneficial for reporters and students who need to extract essential information. Our demo video is available at (\url{https://youtu.be/7ZUBIyqPpbs}). Future work will focus on multilingual support and the integration of multimodal LLMs to handle various data types, such as images and videos, enhancing user control and improving human-computer interaction.

\section{Acknowledgements}
This research was supported by Basic Science Research Program through the National Research Foundation of Korea(NRF) funded by the Ministry of Education(NRF-2022R1C1C1008534), and Institute for Information \& communications Technology Planning \& Evaluation (IITP) through the Korea government (MSIT) under Grant No. 20210-01341 (Artificial Intelligence Graduate School Program, Chung-Ang University).

\bibliography{aaai25}

\newpage


\begin{table}[t!]
\centering
\resizebox{0.9\columnwidth}{!}{%
\begin{tabular}{l|cccccc}
\hline
\textbf{User}      & \#1  & \#2 & \#3 & \#4 & \#5  & \#6  \\ \hline
\textbf{SUS Score} & 92.5 & 90  & 75  & 80  & 92.5 & 87.5 \\ \hline
\end{tabular}
}
\caption{Result of SUS scores. Each participant's score was calculated based on their responses to the ten SUS questions.}
\label{tab-sus}
\end{table}

\section{Details for Evaluations}

We conducted two user studies to evaluate \textsc{SummPilot} using HCI questionnaires. The first study focused on system usability, examining interactive components and summary quality. The second compared user preferences and information acquisition efficiency between \textsc{Basic Mode} and \textsc{Advanced Mode}. We used the think-aloud approach~\cite{van1994think} to capture users' thoughts, gaining insights into their comprehension and intentions. Each 45-minute session was conducted via Zoom.

\subsection{Participants and Instructions}
Following system usability testing principles~\cite{nielsen1994usability}, which suggest six evaluators are sufficient for prototype evaluation, we recruited six participants from volunteer students in our department. These participants had diverse backgrounds, including various  undergraduate majors, ages, and genders. Before experiment, participants received instructions simulating real-world scenarios as system users, with guidance tailored to the system's primary objectives.

\setlength{\parindent}{0cm}{
\begin{minipage}[t]
{\linewidth}\raggedright
\setlength{\parindent}{0cm}{
\hrule

{\small As an intern reporter, your task is to analyze the assigned news topic and write a summary of it. To accomplish this, you will receive multiple news articles covering a specific news topic. You can use an application to assist you with this task. Your performance will be evaluated based on two criteria: 1) your understanding of the news topic and 2) the effort you put into writing the summary.}
\hrule
}
\end{minipage}
}

\subsection{System Usability Test}
\label{sec-exp-system}

\subsubsection{Setup}
\label{sec-exp-system-setup}
Each participant was randomly assigned two documents from Multi-News~\cite{fabbri-etal-2019-multi}, which they reviewed according to the provided instructions. After interacting with the system, participants evaluated usability using two questionnaires: the system usability scale (SUS)~\cite{jordan1996usability} to measure overall usability and a custom-designed usefulness questionnaire to assess how well each interface component facilitated document understanding and summary customization.

\begin{table}[t!]
\centering
\resizebox{1.0\columnwidth}{!}{
\begin{tabular}{lc}
\hline
\textbf{SUS Questionnaire}         & \textbf{SUS Score} \\ \hline \hline
1. I think that I would like to use this system frequently. & 4.5 (0.5) \\ 
2. I found the system unnecessarily. & 1.3 (0.5) \\ 
3. I thought the system was easy to use.  & 4.7 (0.5)  \\ 
4. I think that I would need the support of a technical & 2.0 (1.1) \\
\quad person to be able to use this system. & \\ 
5. I found the various functions in this system were  & 4.7 (0.4) \\ 
\quad well integrated. & \\ 
6. I thought there was too much inconsistency in this  & 1.2 (0.4)  \\ 
\quad system.  & 1.2 (0.4)  \\ 
7. I would imagine that most people would learn to use  & 4.2 (0.6) \\ 
\quad this system very quickly.  &      \\ 
8. I found the system very cumbersome to use.  & 1.7 (0.5) \\ 
9. I felt very confident using the system. & 4.5 (0.5) \\ 
10. I needed to learn a lot of things before I could get  & 1.9(0.4) \\
\quad \, going with this system.  &  \\
\hline \end{tabular}
}
\caption{Detailed SUS score. Each score is reported as average (standard deviation).}
\label{tab-sus-detail}
\end{table}

\paragraph{SUS Questionnaire \& Result}\label{app-sus-q} The SUS is a 10-item scale for overall subjective assessment. The SUS score is calculated as $2.5 * \textit{Total SUS contribution}$. For questions 1, 3, 5, 7, and 9, the contribution is calculated as $\textit{Score} - 1$ (higher scores indicate better usability). For questions 2, 4, 6, 8, and 10, it is calculated as $5 - \textit{Score}$ (lower score is preferable). Table~\ref{tab-sus} shows the SUS results for the six participants, with an average score of 86.3, classifying it as "excellent" based on established thresholds~\cite{will2021measuring}. Each question and its corresponding score are shown in Table~\ref{tab-sus-detail}.

\paragraph{Usefulness Questionnaire \& Result}\label{app-custom} The usefulness questionnaire consists of nine items: five evaluating interaction elements, such as the graph and relation extraction, and four assessing the quality of the generated summary. Table~\ref{tab-custom-detail} shows that participants rated the semantic graph highly for document understanding, with an average score of 4.4, and were fully satisfied with the intuitive interaction through relation extraction, which received an average score of 5.0. Participants highlighted the usefulness of the possible errors feature in evaluating the generated summary. These results highlight \textsc{SummPilot}'s strengths in enhancing document comprehension and supporting interactive summarization and evaluation.

\begin{table}[t!]
\centering
\resizebox{1.0\columnwidth}{!}{
\begin{tabular}{lc}
\hline
\textbf{Usefulness Questionnaire}                       & \textbf{Score} \\ \hline \hline
Graph helps to understand the content of the given  & 4.4 (0.8)  \\ 
article. &  \\ 
Entity Clustering helps to understand the content  & 3.8 (0.7)      \\ 
of the given article. &  \\ 
Relation Extraction table is intuitive to include or & 5.0 (0.0)      \\ 
 exclude specific relationships. &  \\ 
It is easy to control the generated summary to    & 4.6 (0.5)  \\
meet my requirements.    &  \\
Possible Error helps to evaluate the generated  & 4.6 (0.5) \\
summary. & \\
\hline
The generated summary effectively includes the  & 4.8 (0.4)  \\ 
information I requested (``\texttt{o}''). &    \\ 
The generated summary effectively excludes the  & 4.8 (0.4)      \\ 
information I requested (``×'' ).   &  \\ 
I feel satisfied with the generated summary.  & 4.8 (0.4)      \\ 
I feel that the generated summary is trustworthy. & 4.4 (0.8)      \\ \hline
\end{tabular}
}

\caption{Detailed usefulness score. Each score is reported as average (standard deviation).}
\label{tab-custom-detail}
\end{table}

\subsection{Comparative Usefulness Test}
\label{sec-exp-comparative}

\subsubsection{Setup} 

To compare the effects and usefulness of \textsc{Basic Mode} and \textsc{Advanced Mode}, we conducted experiments with six new participants.  Each participant summarized two randomly selected articles from Multi-News \cite{fabbri-etal-2019-multi} using each mode in different sequences. Afterward, participants completed the standard USE questionnaire~\cite{lund2001measuring}. Following SummHelper~\cite{slobodkin-etal-2023-summhelper}, participants rated their preference for \textsc{Basic Mode} (closer to 1) and \textsc{Advanced Mode} (closer to 5). Additionally, two custom items for system's primary objectives were included. Detailed questions are provided in Table~\ref{tab-use-detail}.

To further assess the impact of the \textsc{Advanced Mode} on document comprehension, we conducted an evaluation using problem-solving tasks. To minimize the influence of participants' prior knowledge, we presented them with hypothetical news articles accompanied by true/false questions, ensuring they had no prior familiarity with the content. These virtual articles were generated using ChatGPT\footnote{\url{http://chat.openai.com}}. Participants completed the tasks using different modes for each document, and we measured both their speed and accuracy.

\begin{table}[t!]
\centering
\resizebox{0.95\columnwidth}{!}{
\begin{tabular}{l}
\hline
\textbf{USE Questionnaire} \\ \hline \hline
It helps me be more effective. \\ 
It helps me be more productive.\\ 
It is useful.                   \\ 
It gives me more control over output.\\ 
It makes it easier to achieve the desired output.\\ 
It saves me time when I use it. \\ 
It meets my needs in addressing the task.\\ 
It does everything I would expect it to do.\\ \hline
It is easy to use. \\ 
It is simple to use.  \\
It is user-friendly. \\ 
It requires the fewest steps possible to accomplish the task. \\ 
It is flexible.  \\ 
Using it is effortless.\\ 
I can use it without written instructions.\\ 
I don’t notice any inconsistencies as I use it. \\ 
Both occasional and regular users would like it.\\ 
I can recover from mistakes quickly and easily.\\ 
I can use it successfully every time. \\ \hline
I learned to use it quickly. \\
I easily remember how to use it. \\ 
It is easy to learn to use it.   \\ 
I quickly became skillful with it. \\ \hline

I would recommend it to a friend.\\ 
It is fun to use. \\ 
It works the way I want it to work.\\ 
It is wonderful.\\ 
I feel I need to have it.  \\ 
It is pleasant to use.  \\
\hline
It helps me understand the content of the given article.* \\ 
\hline
It helps me adjust the generated summary.*  \\
\hline
\end{tabular}
}
\caption{Detailed USE questionnaire. Two additional custom questions are marked with (*).}
\label{tab-use-detail}
\end{table}

\begin{table}[t!]
\centering
\resizebox{0.75\columnwidth}{!}{%
\begin{tabular}{lc}
\hline
\textbf{Dimension}                           & \textbf{Score} \\ \hline\hline
Usefulness                                   & 4.3 (0.3)      \\
Ease of Use                                  & 2.9 (0.3)      \\
Ease of Learning                             & 2.6 (0.1)      \\
Satisfaction                                 & 4.5 (0.3)      \\
Understanding the Article\textsuperscript{*} & 4.3 (0.8)      \\

Adjusting the Summary\textsuperscript{*}     & 4.8 (0.4)      \\ \hline
\end{tabular}
}
\caption{Result of USE questionnaire. Two additional custom questions are marked with (*). Scores are reported as averages (standard deviations).}
\label{tab-use}
\end{table}

\begin{table}[t!]
\centering
\resizebox{0.9\columnwidth}{!}{%
\begin{tabular}{l|cc}
\hline
                 & \textbf{\textsc{Basic Mode}} & \textbf{\textsc{Advanced Mode}} \\ \hline\hline
Correct Rate     & 90\%                 & 90\%                   \\
Time Spent & 3:33                 & 5:02                   \\ \hline
\end{tabular}
}
\caption{Comparison of problem-solving results.}
\label{tab-prob}
\end{table}

\subsubsection{Result}

Table~\ref{tab-use} presents the USE questionnaire results. \textsc{Advanced Mode} received higher scores in ``Usefulness'' (4.3) and ``Satisfaction'' (4.5). Users preferred \textsc{Advanced Mode} for ``Understanding the Article'' (4.3) and ``Adjusting the Summary'' (4.8), emphasizing its effectiveness in facilitating comprehension and content customization. Conversely, \textsc{Basic Mode}, known for its simplicity and minimal user intervention, excelled in ``Ease of Use'' (2.9) and ``Ease of Learning'' (2.6). The integration of both modes in \textsc{SummPilot} offers flexibility to accommodate diverse user preferences and objectives, demonstrating its adaptability in balancing simplicity with interactive customization features.

To further validate the effectiveness of \textsc{Advanced Mode} and its components, we compared it with \textsc{Basic Mode} in problem-solving tasks. Participants solved problems under two conditions: 1) using \textsc{Advanced Mode} with interactive features, and 2) using \textsc{Basic Mode} with automated summaries. The results, shown in Table~\ref{tab-prob}, indicate that the interactive features in \textsc{Advanced Mode}—such as the semantic graph, relation extraction, and entity clustering—significantly enhanced users' document understanding and facilitated faster task completion compared to automated summaries. These findings align with the positive feedback observed in Table~\ref{tab-custom-detail}, emphasizing that the interactive components are not only intuitive but also efficient in supporting rapid comprehension.

\section{System Details}
Our system's interface is designed using the Bootstrap\footnote{\url{https://github.com/twbs/bootstrap}} framework. For the backend, we utilized Python's Flask\footnote{\url{https://github.com/pallets/flask}} framework, with each module leveraging OpenAI's \texttt{GPT-4o}. We note that \textsc{SummPilot} can be deployed using only a CPU.

\newpage

\section{Prompts}
\label{app-prompt}

\subsection{Prompt for Automatic Summarization and Interactive Summarization}

\begin{minipage}[t]
{\linewidth}\raggedright
\setlength{\parindent}{0cm}
\hrule

\small{
\texttt{\textcolor{TealBlue}{System:}} You are a helpful assistant. Your job is to summarize given documents.

\texttt{\textcolor{TealBlue}{User:}} Please summarize multiple documents into a single, concise document. \\ Exclude any unrelated noisy content, such as advertisements. The input documents will be provided next. \\ Format the output as follows: \newline
[Summary]\\ Content  

\texttt{\textcolor{TealBlue}{User:}} \newline
[Article 1] \newline
\textcolor{gray}{\# Content of article 1} \newline
[Article 2] \newline 
\textcolor{gray}{\# Content of article 2} \newline
\texttt{\textcolor{TealBlue}{Assistant:}} \newline
[Summary] \newline
\textcolor{gray}{\# Initial summary} \newline
\texttt{\textcolor{TealBlue}{User:}} Thank you. Please address the following requests in your summary. \newline
* Ensure that the summary includes content related to the triple subject-relation-object.\newline
* Remove any content related to the triple subject-relation-object.\newline
* Format the output as follows: \newline
[Summary]\newline
Content
}
\hrule
\end{minipage}

\subsection{Prompt for Factual Decomposition}

{
\setlength{\parindent}{0cm}
\begin{minipage}[t]
{\linewidth}\raggedright

\hrule
{\small
\texttt{\textcolor{TealBlue}{System:}} You are a helpful assistant. Your job is to break down given sentences into independent facts.\\
Do not generate chat-style responses such as ``Sure!'', ``Here's the output''. Do not generate overly repetitive facts.\\

\texttt{\textcolor{TealBlue}{User:}} Please breakdown the following sentence into independent facts:\\
He is a producer and engineer, having worked with a wide variety of artists, including Willie Nelson and Taylor Swift. \\

\texttt{\textcolor{TealBlue}{Assistant:}} \\
* He is a producer.\\
* He is an engineer.\\
* He has worked with a wide variety of artists.\\
* Willie Nelson is an artist.\\
* He has worked with Willie Nelson.\\
* Taylor Swift is an artist.\\
* He has worked with Taylor Swift.\\

\texttt{\textcolor{TealBlue}{User:}} Please breakdown the following sentence into independent facts:\\
\textcolor{gray}{\# Each sentence from the generated summary to evaluate}
}

\hrule
\end{minipage}
}

This prompt is a slightly modified version of the one suggested by FactScore \cite{min-etal-2023-factscore} to suit our task. Due to space constraints, only one example is included here, but we used few-shot prompting in the actual process.
\subsection{Prompt for Relation Extraction and Mention Clustering}
\begin{minipage}[t]
{\linewidth}\raggedright
\setlength{\parindent}{0cm}
\hrule

{\small
\texttt{\textcolor{TealBlue}{System:}} You are a helpful assistant. Your job is to extract relation triples from given documents.\newline
Do not generate chat-style responses such as ``Sure!'', ``Here's the output''.\newline
\texttt{\textcolor{TealBlue}{User:}} Please extract important and concise relational triples from the document. Identify relationships between each pair of entities.\newline
Each triple consists of a subject, a relation, and an object. \newline
Output the results in the following format: \newline
[Relation Triples]\newline
* $<$Subject$|$Relation$|$Object$>$\newline
* $<$Subject$|$Relation$|$Object$>$\newline
* $<$Subject$|$Relation$|$Object$>$\newline
\texttt{\textcolor{TealBlue}{User:}} \newline
[Document] \newline
\textcolor{gray}{\# Content of input document}

\texttt{\textcolor{TealBlue}{Assistant:}} \newline
[Relation triples] \newline
\textcolor{gray}{\# Extracted triples}

\texttt{\textcolor{TealBlue}{User:}} Thank you. Please perform coreference resolution on the entities extracted from the triple. An entity includes a subject and an object, not a relation.

* Coreference resolution is the task of identifying and linking different noun phrases that represent the same entity.

* Group together multiple noun phrases that represent the same entity into one triple.

* For example, 'Los Angeles' and 'L.A.' should be grouped together since they represent same entity. 'Jordan' and 'Michael Jordan' should be reoupted together if they represent the same entity.

* Read the given document, identify coreferences for the subject and object of the extracted triple, and combine them into a single triple.

* For example, if you have $<$Los Angeles$|$Located in$|$California$>$ in the generated triple list and L.A. in the document, combine them into $<$[Los Angeles+L.A.]$|$Located in$|$California$>$.

* If you can't find any coreference, leave the triple as it is. Print the original triple in this case.

* Output the results in the following format: \newline
[Coreference Resolution] 

* $<$[Subject1a+Subject1b]$|$Relation1$|$Object1$>$

* $<$[Subject2a+Subject2b+Subject2c]$|$Relation2$|$Object2$>$
}
\hrule
\end{minipage}


\subsection{Prompt for Measuring Factual Consistency}

\begin{minipage}[t]
{\linewidth}\raggedright
\setlength{\parindent}{0cm}
\hrule
{\small
\texttt{\textcolor{TealBlue}{System:}} You are a helpful assistant. Your job is to judge if each statement is true or false based on the given document.\\
Do not generate chat-style responses such as ``Sure!'', ``Here's the output''. Answer only in the format ``True'' or ``False''.\\

\texttt{\textcolor{TealBlue}{User:}} \\

[Document]\\ \textcolor{gray}{\# Given input documents}\\

\texttt{\textcolor{TealBlue}{User:}} Please judge if the following statement is correct based on the given document:\\ \textcolor{gray}{\# Each atomic fact to evaluate}
}
\vspace{0.5mm}
\hrule
\end{minipage}


\end{document}